\documentclass[runningheads]{llncs}
\usepackage{float}
\usepackage{amsmath}
\usepackage[T1]{fontenc}
\usepackage{multirow}
\usepackage{graphicx}
\usepackage{graphicx}
\usepackage{color}
\usepackage{amsmath}
\usepackage[backend=biber,style=numeric, sorting=none]{biblatex}
\begin{document}
\title{Brain Tumor Detection through Thermal Imaging and MobileNET}
\titlerunning{Brain Tumor through TI and MN}
%
\author{Roham Maiti\inst{1}\orcidID{0009-0005-1010-4107} 
Debasmita Bhoumik\inst{1}\orcidID{0000-0002-1795-5298} }

\authorrunning{R. Maiti, D. Bhoumik}
%
\institute{
Sister Nivedita University, Kolkata, India \\
\email{maitiroham2003@gmail.com, debasmita.b@snuniv.ac.in}
}

\maketitle              
\begin{abstract}
Brain plays a crucial role in regulating body functions and cognitive processes, with brain tumors posing significant risks to human health. Precise and prompt detection is a key factor in proper treatment and better patient outcomes. Traditional methods for detecting brain tumors, that include biopsies, MRI, and CT scans often face challenges due to their high costs and the need for specialized medical expertise. Recent developments in machine learning (ML) and deep learning (DL) has exhibited strong capabilities in automating the identification and categorization of brain tumors from medical images, especially MRI scans. However, these classical ML models have limitations, such as high computational demands, the need for large datasets, and long training times, which hinder their accessibility and efficiency. Our research uses MobileNET model for efficient detection of these tumors. The novelty of this project lies in building an accurate tumor detection model which use less computing re-sources and runs in less time followed by efficient decision making through the use of image processing technique for accurate results. The suggested method attained an average accuracy of 98.5\%.

\keywords{Brain Tumor  \and CNN \and MobileNET \and Image Processing \and Tumor Detection \and Thermal Imaging}
\end{abstract}
\section{Introduction}
As the primary nervous system organ, the brain regulates physiological functions, interprets sensory data, and supports memory, emotion, and cognition \cite{incir2024improving}. Composed of billions of neurons, it governs essential activities such as movement, speech, and vital functions like breathing and heart rate, while enabling complex abilities like problem-solving, creativity, and reasoning. Its role in maintaining both physical and mental health is fundamental and irreplaceable.

\vspace{10pt}

\noindent An abnormal cell proliferation inside the brain that interferes with its highly ordered structure and functions is referred to as a brain tumour\cite{miles1976british}. Even non-malignant tumors can lead to serious neurological issues by pressing against surrounding brain structures. Signs and symptoms of brain tumors are not uniform and may vary significantly, based on their size and location, ranging from headaches and seizures to cognitive difficulties and vision changes. Left untreated, brain tumors can severely impact general health, potentially leading to life-threatening complications. Early detection and timely intervention are crucial to improving outcomes and preserving quality of life. Brain tumor damage can occur due to increased intracranial pressure, displacement of brain structures, and invasion of healthy tissues \cite{louis20072007,abolanle2020brain}. The World Health Organization (WHO) provides a four-layer classification technique to assist in radiological diagnosis, including integrated diagnosis, histological classification, WHO grading, and molecular information \cite{johnson20172016}.

\vspace{10pt}

\noindent The likelihood of developing brain and central nervous system tumors is influenced by gender and racial disparities. With male-to-female risk ratios ranging from 1.22 to 1.67 for various tumour types, men are typically at a higher risk than women. Racial disparities indicate that White individuals have the highest risk across most tumor types, followed by Black and Other groups. For example, glioblastoma risk is 4.07 in men compared to 2.55 in women \cite{vovoras2014epidemiology}.

\vspace{10pt}

\noindent Conventional methods for diagnosing brain tumors are biopsies, magnetic resonance imaging and computed tomography. While biopsies provide detailed cellular-level information, they are invasive and carry risks. Despite being non-invasive, imaging methods, involve high resource consumption and specialized expertise, restricting their availability in resource-limited environments. Moreover, these methods are expensive and not always practical for widespread use in underdeveloped regions \cite{shankar2017liquid,ostertag1980stereotactic}.

\vspace{10pt}

\noindent As artificial intelligence continues to evolve, machine learning is increasingly recognized for its potential in identifying brain tumors. Unlike conventional methods, ML models are non-invasive, rely on medical imaging, and provide high accuracy when properly trained. Studies have explored models like CNNs for automated brain tumor identification and categorization., achieving high accuracy with architectures like ResNet, MobileNet, and DenseNet \cite{arfan2021classification,sadad2021brain,abdusalomov2023brain}. Researchers have also explored thermal imaging, noise reduction, and innovative preprocessing techniques for enhanced tumor localization and reduced computational costs \cite{KATEB2009T154,8599180,7943218}. Despite these advancements, challenges like reliance on large datasets and high computational requirements persist.

\vspace{10pt}

\noindent \textbf{Novelty of the Proposed Work:}

\noindent This study suggests a number of innovative approaches aimed at increasing accessibility, reducing expenses, and boosting diagnosis accuracy  to address the shortcomings of present diagnostic methods as well as research gaps. The goal is to develop a solution that reduces reliance on costly equipment and intensive computational power, thereby enabling effective tumor detection in resource-constrained environments. By using lightweight machine learning architectures with simulated thermal imaging and refined preprocessing methods, the proposed system addresses the challenges associated with both traditional diagnostic tools and modern ML approaches, providing a reliable and practical alternative for accurate early detection of brain tumors.

\begin{itemize}
\item Integration of lightweight models like MobileNET with thermal imaging techniques for cost-effective brain tumor detection.
\item Development of an OpenCV-based solution to emulate thermal imaging without requiring expensive thermal cameras.
\item Introduction of image enhancement techniques such as blurring and edge detection to improve tumor localization accuracy.
\item Accessibility-focused design, ensuring usability in low-resource settings and for users with limited technical expertise.
\item Demonstration of superior performance, reaching a validation accuracy of 98.8\%, exceeding cutting-edge models like DenseNET and GoogleNET while maintaining low runtime and validation loss.
\end{itemize}

\noindent This paper introduces a system leveraging MobileNET and OpenCV for accurate, cost-effective, and non-invasive brain tumor detection. By lowering processing demands and enhancing accessibility for contexts with limited resources, the suggested framework aims to solve current issues.  Additionally, the system's flexibility guarantees that it may be used on mobile devices, opening up advanced diagnostic methods to a larger audience. The model’s robustness was validated with unseen data, demonstrating correct predictions and reliable tumor localization using advanced image processing techniques.

\section{Literature Review}
Recently, brain tumor detection has garnered significant attention due to the critical role it plays in improving patient outcomes through early diagnosis. Convolutional Neural Networks (CNNs), one of the most sophisticated deep learning models, have been used in a number of studies to automatically identify and categorise brain tumours from MRI scans.  With models trained on balanced datasets demonstrating great accuracy in differentiating between tumour and non-tumor images, architectures such as VGG, ResNET, and MobileNET have demonstrated encouraging outcomes \cite{anjum2025novel}. Also, many types of cameras, along with image processing techniques, were introduced to identify the growth and location of tumors. Many novel brain tumour detection tools are also emerging.  Several brain tumour detection and classification investigations have been published.

\vspace{5pt}

\noindent Quy Thanh Lu et al. \cite{lu2024improving} made an automated brain tumor diagnosis system using deep transfer learning and fine-tuning on MRI images. Utilizing the MobileNet model, 3264 subjects' data (926 glioma, 937 meningioma, 901 pituitary tumors, and 500 normal) yielded high accuracy rates: 97.24 percent validation, 97.86 percent test, and 97.85 percent F1 score in four-class classification, and 100 percent in two-class classification. These results demonstrate a robust machine learning tool to aid doctors in accurate and efficient brain tumor diagnosis.

\vspace{5pt}

\noindent Saeedi et al. \cite{saeedi2023mri} proposed a study which utilized a dataset of 3,264  brain MRI, including meningioma, pituitary tumors, glioma, and healthy cases, to develop a new 2D CNN alongside a convolutional auto-encoder  for brain tumor detection. Following preprocessing and augmentation, a modified auto-encoder network and the 2D CNN—which consists of eight convolutional and four pooling layers—were trained and evaluated. The 2D CNN and convolutional autoencoder demonstrated strong performance, achieving training accuracies of approximately 96\% and 95\%, respectively, with ROC AUC scores close to 1, which indicates excellent classification capability.
Additionally, ML classifiers were evaluated, with K-Nearest Neighbors attaining the highest accuracy at 86 percent. Statistical analysis confirmed significant differences in performance (p < 0.05) between the developed networks and conventional machine learning methods.

\vspace{5pt}

\noindent Swarup et al. \cite{swarup2023brain} made an automatic detection of brain tumors in MRI is crucial for diagnosing neurological disorders. This study suggests a deep CNN model for MRI brain tumour detection, that is a typical application for CNNs.  Following preprocessing, the model compares the performance of GoogleNet and AlexNet architectures using metrics including accuracy, sensitivity, specificity, and AUC. GoogleNet achieved higher accuracy (99.45 percent) and sensitivity (99.75 percent) than AlexNet, which had an accuracy of 98.95 percent and sensitivity of 98.4 percent, while also using fewer parameters. Given its high accuracy and efficiency, the model offers strong potential as a support tool for radiologists in clinical diagnoses.

\vspace{5pt}

\noindent Bousselham et al. \cite{bousselham2018brain} present a method to extract the effects of the temperature of brain tumors from MRI. MRI, a modality rich in tissue information based on nuclear magnetic resonance parameters, reveals that tumor cells generate more heat than healthy cells, affecting these parameters—particularly the spin-lattice relaxation time, T1. The temperature distribution had been computed by employing the Pennes BioHeat Transfer Equation, applied to a head model comprising three concentric spheres and a spherical tumor, and solved via the Finite Difference Method.

\vspace{5pt}

\noindent Yoshihiro Tanaka et al. \cite{tanaka2010development} created a biocompatible real-time tactile sensing system to improve intraoperative brain tumor diagnosis. The active sensor, using balloon expansion, measures tactile properties like softness and smoothness, with potential applications in distinguishing tumors from healthy brain tissue. Tests on soft samples with varied stiffness and surface conditions demonstrated the sensor’s ability to differentiate materials, though performance is affected by boundary conditions. Initial measurements on porcine brain tissue successfully discriminated between white and gray matter, suggesting promising application for brain tumor detection.

\vspace{5pt}

\noindent The work by Amran Hossain et al. \cite{hossain2023brain} presents MSegNet, a lightweight model for segmenting brain tumors in reconstructed microwave images, and BINet, a classification network for the same image type. Using a dataset of 300 RMW brain images augmented to 6000 per fold, they evaluated the models via 5-fold cross-validation. MSegNet attained an IoU of around 87\% and a Dice score of approximately 93\%, while BINet showed strong classification performance with about 98\% accuracy and 99\% specificity on segmented images. These results support the sensors-based microwave brain imaging system’s application for efficient tumor segmentation and classification.

\section{Background}
\noindent Advances in machine learning and imaging techniques have significantly enhanced brain tumor detection. MobileNET and thermal imaging are pivotal in improving diagnostic accuracy and efficiency.

\subsection{MobileNET}

\begin{figure}[H]
\begin{center}
\includegraphics[width=1\textwidth, height=0.35\textheight]{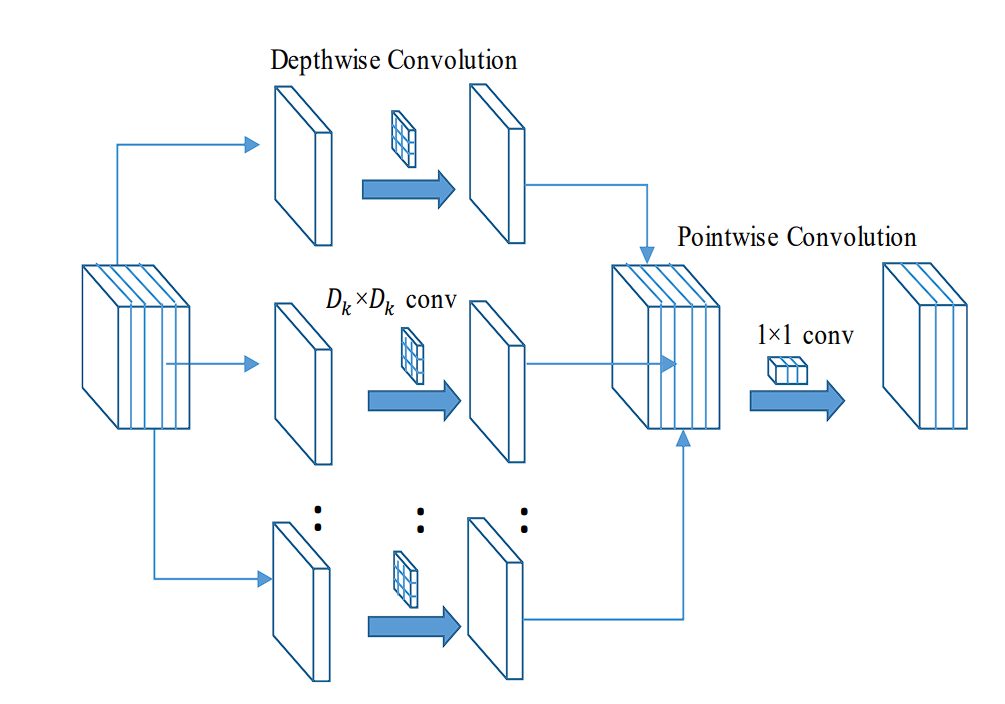}
\caption{Fig: Functioning of a MobileNET detection model}
\label{fig:mobilenet}
\end{center}
\end{figure}

This architecture, which uses depthwise separable convolutions to minimise computational complexity in resource-constrained applications, divides the standard convolution into a depthwise convolution and a pointwise one.

- \textbf{Depthwise Convolution:}
  \begin{itemize}
      \item Uses one filter per input channel.
      \item Decreases complexity from  
\( K \times K \times C \times D \times H \times W \)  
to  
\( K \times K \times C \times H \times W \),  
 here \( K \) represents the kernel size,  
\( C \) signifies the number of input channels,  
\( D \)  represents the number of output channels, and  
\( H, W \) denotes to the spatial dimensions.

  \end{itemize}

- \textbf{Pointwise Convolution:}
  \begin{itemize}
      \item Combines channel features and adjusts the number of output channels.
      \item Enables learning complex inter-channel patterns.
  \end{itemize}

After convolution layers, a global average pooling (GAP) layer reduces parameters by averaging spatial dimensions:
\[
\text{GAP}_c = \frac{1}{H \times W} \sum_{i=1}^{H} \sum_{j=1}^{W} f_{ijc},
\]
where \( f_{ijc} \) 
  indicates the activation value at position \( (i, j) \) in channel \( c \), and \( H \) and \( W \) refer to the spatial dimensions—height and width—of the feature map.

In contrast to heavier CNN architectures like VGG16 or ResNet50, MobileNet offers a balance between accuracy and computational efficiency. Its smaller footprint allows it to run on devices with limited memory and processing power, making it highly suitable for remote or rural healthcare applications where diagnostic infrastructure is minimal.

\subsection{Thermal Imaging}  
Thermal imaging identifies temperature variations caused by tumors' metabolic activity. Infrared cameras capture these differences, aiding early diagnosis. OpenCV’s \texttt{applyColorMap} function was used with \texttt{COLORMAP\_JET} to visualize grayscale images with pseudocolor for better interpretation.

\subsection{Gaussian Blurring}  
Gaussian blurring smooths images by convolving them with a Gaussian function:
\[
\Gamma(x, y) = \frac{1}{2 \pi \sigma^2} \exp{\left(-\frac{x^2 + y^2}{2 \sigma^2}\right)},
\]
where \( \sigma \) determines the intensity of the blur, and \( (x, y) \) represent the pixel distances from the origin.

\subsection{Canny Boundary Detection:}
Boundary detection locates object boundaries in an image by identifying intensity changes. The Canny algorithm involves:

\noindent 1. \textbf{Gaussian Smoothing:} The image \( I(x, y) \) undergoes smoothing using a Gaussian filter \( \Gamma(x, y, \sigma) \):
   \[
   \Gamma(x, y, \sigma) = \frac{1}{2\pi\sigma^2} \exp\left(-\frac{x^2 + y^2}{2\sigma^2}\right)
   \]
   Smoothed image:
   \[
   I_s(x, y) = I(x, y) * \Gamma(x, y, \sigma)
   \]

\noindent 2. \textbf{Gradient Calculation:} Gradients \( \Gamma_x(x, y) \) and \( \Gamma_y(x, y) \), along with the magnitude \( \Gamma(x, y) \) and direction \( \theta(x, y) \):

   \[
   \Gamma(x, y) = \sqrt{\Gamma_x(x, y)^2 + \Gamma_y(x, y)^2}, \quad \theta(x, y) = \arctan\left(\frac{\Gamma_y(x, y)}{\Gamma_x(x, y)}\right)
   \]

\noindent 3. \textbf{Non-Maximum Suppression:} Retain local maxima along the gradient direction.

\noindent 4. \textbf{Double Thresholding:} Classify pixels using thresholds \( T_{\text{high}} \) and \( T_{\text{low}} \):
   \[
   \text{Strong edges: } \Gamma(x, y) \geq T_{\text{high}}, \quad \text{Weak edges: } T_{\text{low}} \leq \Gamma(x, y) < T_{\text{high}}
   \]

\noindent 5. \textbf{Edge Tracking by Hysteresis:}  Involves linking weak edges to adjacent strong edges to complete the edge map.

\section{Proposed Model}
\begin{figure}[H] 
\includegraphics[width=\textwidth]{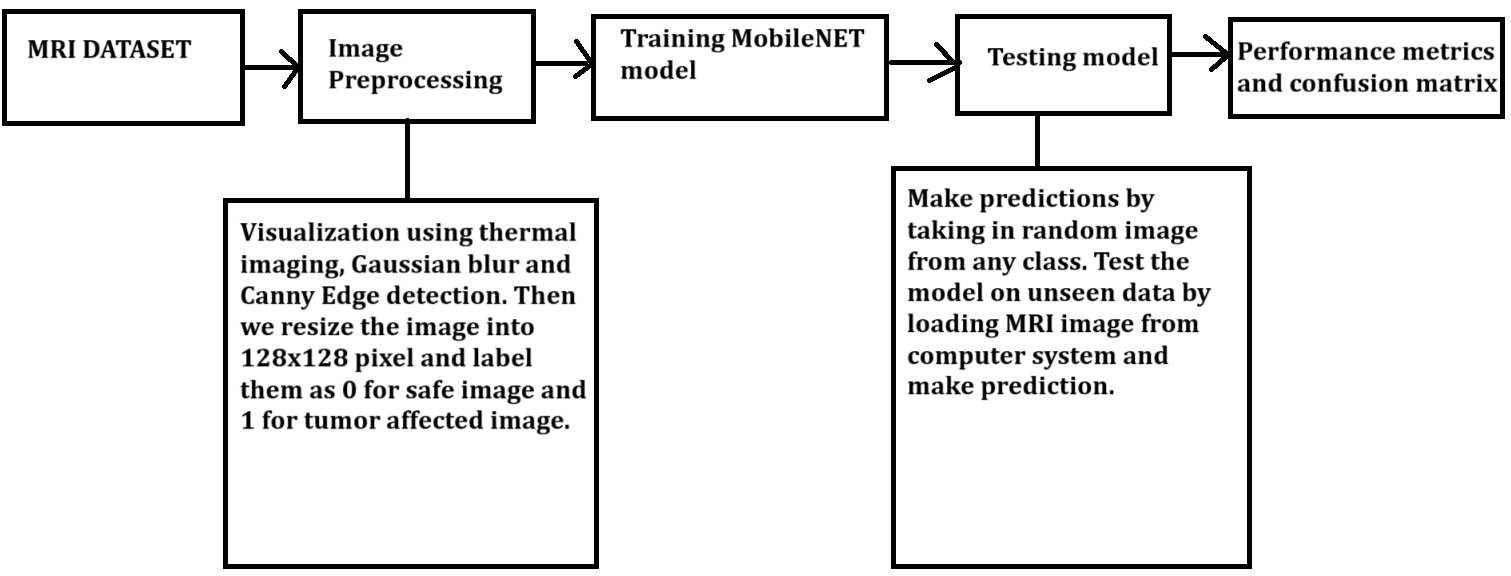}
\caption{Working of the system} \label{fig:workflow}
\end{figure}

The workflow is given in fig \ref{fig:workflow}.

\vspace{5pt}

\noindent Dataset is made by taking a portion of MRI  from an existing Kaggle dataset\cite{msoud_nickparvar_2021} and loading it into Kaggle. The images are then loaded into Google Colab notebook. Necessary Python libraries like numpy, matplotlib, pandas, seaborn, random, os scikit-learn and tensorflow are also imported. 

All MRI images used were anonymized and publicly available through open datasets. Ethical usage of medical data remains a core focus, and future versions of the project will explore integrating privacy-preserving techniques such as differential privacy to ensure compliance with global health data regulations.

\vspace{5pt}

\noindent After this data visualization is done by applying thermal imaging, Gaussian blur and Canny edge detection. This steps helps us to set parameters such as kernel size, standard deviation and type of colormap that is to be used in image processing during later steps.

\vspace{5pt}

\noindent We process the image by labeling as 0 for safe images and 1 for tumor affected images. We read the image in graysacle mode and reshape the images into 128x128 pixels and store the processed label and images. After that, we scale the values between 0 and 1 and normalise the pixel values by dividing them by 255. We split the data into 80\% for training and 20\% for testing. The data is reshaped to match the input shape required by deep learning models. Then convert the labels into one-hot encoded format. Grayscale images are converted into RGB images as MobileNET model has 3 input channels.

\vspace{5pt}

\noindent The MobileNET model by is made by initializing it with pre-trained weights from ImageNet. Top layers are excluded as we are using binary classification and set the shape of the input images to 128x128 with 3 channels (RGB).There are 13 layers which contains depth-wise and point-wise convolutions along with an averaging layer and 2 dense layers. We use global average pooling to reduce the spatial size of the feature maps.  There are two distinct dense layers, with 256 neurons entirely linked in the second-to-last dense layer.  The final dense layer, which has two neurons and makes predictions using the softmax activation function, receives its output.  MobileNet input and custom classification layers are then used to build the model.  Lastly, to avoid updates during training, the model's layers are frozen.  We will only train the recently introduced layers.  The formula for softmax function is given by-
\[
\text{softmax}(z_i) = \frac{e^{z_i}}{\sum_{j=1}^{n} e^{z_j}}
\]

\begin{itemize}
    \item $z_i$: The $i$-th element of the input vector $z$.
    \item $n$: The total number of elements in the input vector $z$.
    \item $e^{z_i}$: Exponentiation of the $i$-th element to ensure all values are positive.
    \item $\sum_{j=1}^{n} e^{z_j}$: Adding up the exponential outputs of every component in the vector, which acts as a normalizing factor to ensure the output values sum to 1.
\end{itemize}

\vspace{5pt}

\noindent We obtain results which is discussed in the result section. To test system model on unseen data we upload image from computer and process the image as done earlier. We use the model made for prediction purpose. Along with making prediction, the system will simultaneously apply Gaussian blur, thermal imaging and Canny edge detection. The output is displayed in fig \ref{fig:unseen_results}

\section{Result}
\subsection{Dataset}
For our work we have made our dataset, sourced from an existing Kaggle dataset\cite{msoud_nickparvar_2021}. The composition of the data set is as follows- 

\[
\begin{array}{|c|c|c|}
\hline
\textbf{Category} & \textbf{Subcategory} & \textbf{Number of Images} \\
\hline
\textbf{No Tumor} &  & 1500 \\
\hline
\multirow{3}{*}{\textbf{Tumor}} & Glioma & 500 \\
 & Pituitary & 500 \\
 & Meningioma & 500 \\
\hline
\textbf{Total} &  & 3000 \\
\hline
\end{array}
\]

\subsection{Parameters}
\noindent The MobileNet model is trained using a batch size of 32 over 50 epochs. Adam is selected as the optimizer, configured with a learning rate of 0.0001. We used log loss/ categorical cross entropy for monitoring loss. We used early stopping to stop training of model if validation loss do not improve for 5 consecutive epochs.

\vspace{5pt}

\noindent For our research we used a kernel of 5x5, standard deviation of 1 is used for Gaussian blur and we have set the upper and lower threshold as 20 for edge detection. We used COLORMAP\_JET from OpenCV for thermal imaging purpose. We have taken results of 5 compilations and found out the average validation accuracy of the model

\subsection{System configuration}
We have used computer system with intel core i5 having stable internet connection. The entire system is developed on Google Colab as it comes with preloaded libraries needed for this task and for collaboration. We have used T4 GPU for faster computation.

\subsection{Experimental Results} 

\begin{figure}[htb] 
\includegraphics[scale = 0.7]{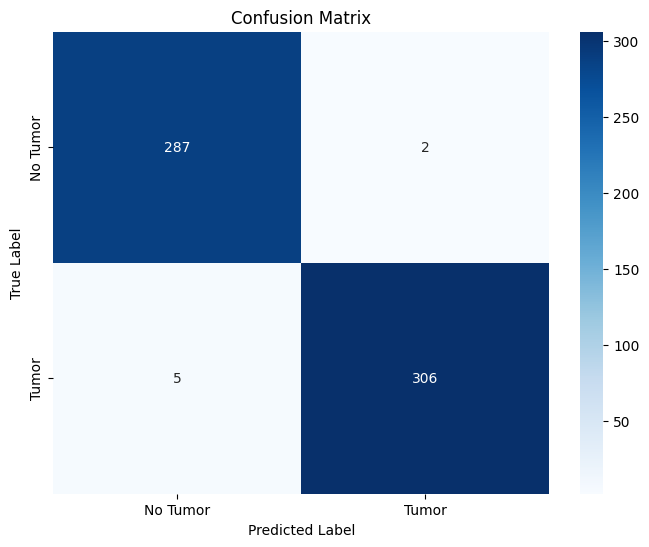}
\caption{Confusion Matrix} \label{fig:confusion_matrix}
\end{figure}

\begin{figure}[H] 
\includegraphics[width=\textwidth]{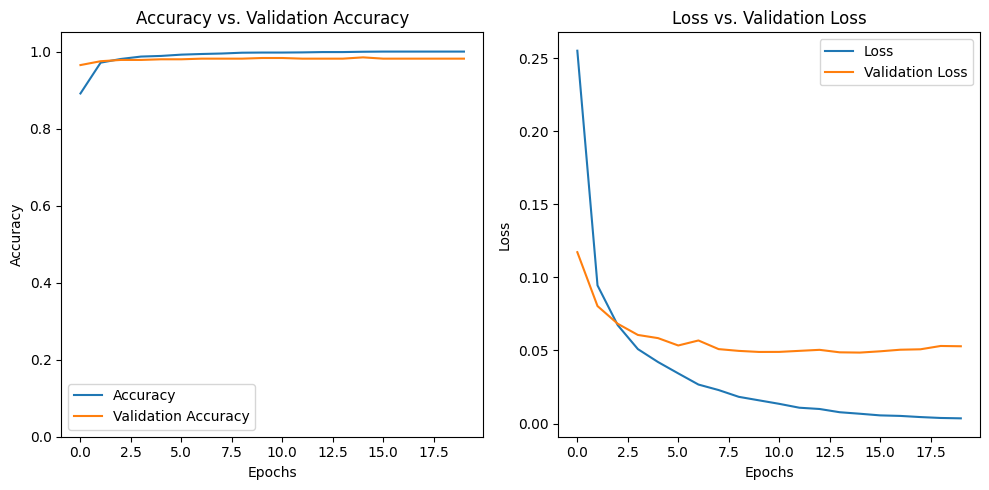}
\caption{Validation Accuracy and Validation Loss} \label{fig:validation_graph}
\end{figure}

A confusion matrix is a helpful tool for assessing model performance in deep learning.  The rows in this matrix reflect real (true) classes, whereas the columns indicate anticipated classes.  Reducing false positives and false negatives is essential for enhancing the model's accuracy and dependability in medical research.  Fig. displays the corresponding confusion matrix.
\ref{fig:confusion_matrix}. The  validation accuracy  and  loss are  shown in Fig \ref{fig:validation_graph}.

\noindent The confusion matrix analysis of the proposed model represents TP (True Positive), TN (True Negative), FP (False Positive), and FN (False Negative) ratio obtained from the testing dataset.\\

$ Precision = \frac{TP}{TP + FP} = \frac{306}{306 + 2} = \frac{306}{308} \approx 0.994 $\\

$ Recall = \frac{TP}{TP+ FN} = \frac{306}{306 + 5} = \frac{306}{311} \approx 0.984 $\\

$ F1\text{-}Score = 2 \times \frac{Precision \times Recall}{Precision + Recall} = 2 \times \frac{0.994 \times 0.984}{0.994 + 0.984} \approx 0.989 $\\

$ Accuracy = \frac{TP + TN}{Total\_Samples} =  \frac{306 + 287}{306 + 287 + 2 + 5} = \frac{593}{600} \approx 0.988 $ \\

\noindent The model achieves a validation accuracy of 98.8\%, surpassing other models while maintaining low runtime and validation loss.

\noindent We tested our model on unseen data by uploading random brain MRI images to check how well the model performs. The results are shown below in Fig \ref{fig:unseen_results}:

\begin{figure}[H] 
\includegraphics[width=\textwidth]{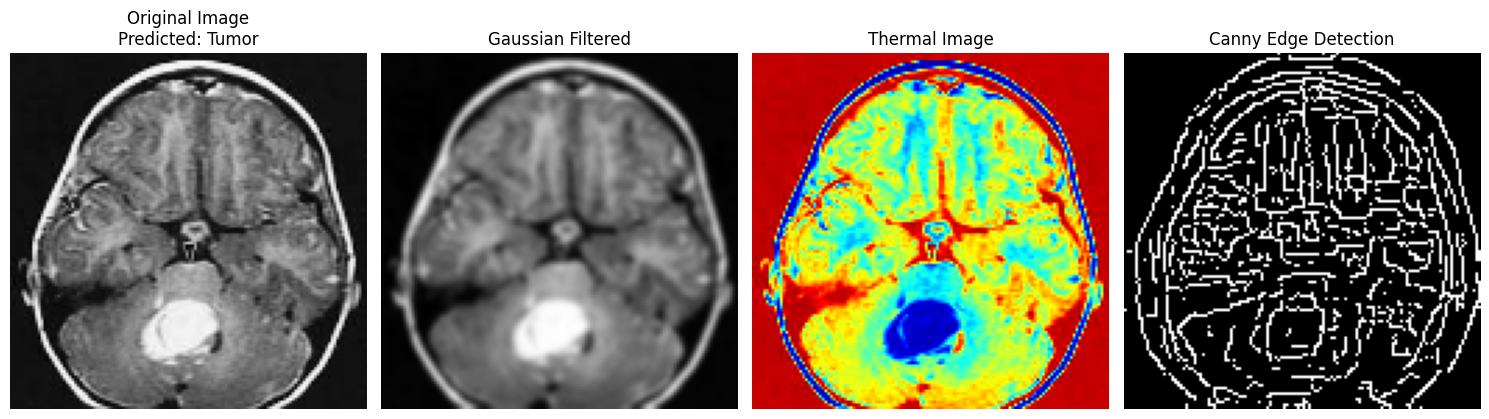}
\caption{Outcome of the Model on Unseen Data} \label{fig:unseen_results}
\end{figure} 

\noindent  As we can see, the model makes correct predictions on unseen data, which helps to ensure its robustness in real-world scenarios. Three types of image processing techniques have been used, namely Gaussian blur, thermal imaging, and Canny edge detection. These techniques assist in detecting the tumor and identifying its location. The image processing helps us to determine weather truly there is any tumor or not, in case of any type of false prediction being done by the model.

While the current model performs strongly on controlled datasets, real-world deployment may face challenges such as variability in MRI resolution, patient movement artifacts, or differing imaging protocols across hospitals. Future iterations of this system could incorporate federated learning to train models across decentralized hospital data without violating patient privacy.

\section{Conclusion}
This work aimed to accurately detect brain tumors and identify affected regions in MRI images by employing image processing techniques and a DL model that achieved over 98\% accuracy with minimal computational resources, making it suitable for resource-constrained devices. Future advancements could involve integrating federated learning to enable collaborative model training across hospitals while preserving patient privacy, enhancing generalizability, and reducing bias. Furthermore, using explainable AI methods like Grad-CAM or LIME would increase openness, aid in the interpretation of model predictions by physicians, and promote confidence in AI-assisted diagnosis.

%
%
%
%

\printbibliography
\end{document}